\documentclass{Interspeech}

\interspeechcameraready

\title{A Novel Data Augmentation Approach for Automatic Speaking Assessment on Opinion Expressions}

\author[affiliation={}]{Chung-Chun}{Wang}
\author[affiliation={}]{Jhen-Ke}{Lin}
\author[affiliation={}]{Hao-Chien}{Lu}
\author[affiliation={}]{Hong-Yun}{Lin}
\author[affiliation={}]{Berlin}{Chen}

\affiliation{Dept. Computer Science and Information Engineering}{National Taiwan Normal University}{Taiwan}
\email{\{takala, jacob, howchien, buffett, berlin\}@ntnu.edu.tw}
\keywords{Automated speaking assessment, Opinion Expression, Data Augmentation}

\usepackage{comment}

\begin{document}

\maketitle

\begin{abstract}
    
    Automated speaking assessment (ASA) on opinion expressions is often hampered by the scarcity of labeled recordings, which restricts prompt diversity and undermines scoring reliability. To address this challenge, we propose a novel training paradigm that leverages a large language models (LLM) to generate diverse responses of a given proficiency level, converts responses into synthesized speech via speaker-aware text-to-speech synthesis, and employs a dynamic importance loss to adaptively reweight training instances based on feature distribution differences between synthesized and real speech. Subsequently, a multimodal large language model integrates aligned textual features with speech signals to predict proficiency scores directly. Experiments conducted on the LTTC dataset show that our approach outperforms methods relying on real data or conventional augmentation, effectively mitigating low-resource constraints and enabling ASA on opinion expressions with cross-modal information.

\end{abstract}

\section{Introduction}
In recent years, technologies for computer-assisted language learning (CALL), such as automated speaking assessment (ASA), have made significant strides to meet the growing demand for scalable and objective evaluation of second-language (L2) speaking proficiency in both academic and professional contexts \cite{eskenazi2009overview, lin2021attention, zechner2008towards}. In L2 speaking examinations, test-takers are often required to express opinions on a given topic or to participate in open-ended discussions, tasks that are crucial for assessing learners’ communicative competence and critical thinking in languages such as English \cite{banno23_slate}. ASA systems that provide immediate feedback and support large-scale scoring are becoming commonplace in both CALL applications and formal testing scenarios.

However, traditional human scoring is time-consuming, labor-intensive, and prone to inconsistency among raters, a problem that is especially acute in high-stakes opinion-expression tasks such as those found in TOEFL, IELTS, and GEPT \cite{lo-etal-2024-effective, qian2019neural, lu2024development}. CALL platforms, through their immediate feedback and interactive instruction, have evolved from early academic prototypes into mature intelligent classroom assistants and robust online testing systems, effectively alleviating many of these challenges. By reducing the burden on human raters and improving scoring consistency, effective ASA for opinion-expression tasks has figured prominently in diverse use cases of language teaching and assessment.

Early ASA approaches combine automatic speech recognition (ASR) with hand-crafted acoustic and linguistic features, such as pronunciation accuracy \cite{chen2010assessment, takai2020deep}, fluency \cite{strik1999automatic}, prosody \cite{coutinho2016assessing},  rhythm \cite{kyriakopoulos2019deep}, and others. These features play a fundamental role in many automated scoring tasks and became the cornerstone of the field \cite{qian2019neural, yu2015using, chen2018end, chen2018automated, park23c_interspeech}. A few pioneering systems extract measures of pronunciation fluency, lexical richness, grammatical accuracy, content relevance, and the like from L2 speech, and then draw on statistical models to predict a composite proficiency score.

More recently, multi-task and multimodal architectures have been introduced to capture complementary information inherent in both the text content and speech signal of a spoken response. The emergence of self-supervised speech encoders (e.g., wav2vec 2.0 \cite{lo-etal-2024-effective, lu2024development, baevski2020wav2vec}), Transformer-based text models \cite{devlin2019bert}, and multimodal large language model (MLLM)  such as Phi-4 multimodal \cite{abouelenin2025phi} gives rise to a variety of end-to-end modeling frameworks that learn response representations jointly from raw audio and textual transcripts, markedly improving scoring accuracy and generalization. Nevertheless, these models still rely excessively on large amounts of annotated training datasets that are frequently scarce for ASA on opinion expressions, thus often leading to overfitting on limited prompts.

To alleviate data scarcity, prior studies explore combining text-to-speech (TTS) techniques with LLM–driven data augmentation \cite{10096105}. These approaches have shown some success on various speech and language processing tasks, such as automatic speech recognition (ASR) \cite{yang2023text, fazel21_interspeech}, emotion recognition \cite{tang23_interspeech}, text classification \cite{kuo2025not, yoo-etal-2021-gpt3mix-leveraging}, and intent recognition \cite{sahu-etal-2022-data}. For the purpose of spoken content generation, these methods typically use LLMs to generate diverse texts, thereby increasing content variety, and subsequently employ TTS systems to synthesize speech from the generated texts, enriching speaker characteristics, prosody, and content. Because synthesized and real texts and audio waves differ in their acoustic feature distributions, there is good reason to apply dynamic instance reweighting or metric-learning techniques to adaptively adjust the importance of each synthesized instance based on its proximity to real counterparts in a embedding space, thus narrowing the domain gap \cite{kuo2025not}. However, to our knowledge, there is still a dearth of work investigating the synergistic effect of combining the aforementioned methods and their extensions to automated scoring of spoken opinion expressions.

In view of this, we present a unified framework for spoken opinion‐expression assessment under the constraint of limited resources, which combines generative data augmentation with neural multimodal modeling. First, we develop a two‐step augmentation pipeline in which an LLM generates proficiency‐conditioned text responses \cite{wang-etal-2023-self-instruct} and a high‐fidelity TTS system synthesizes these text responses into spoken utterances while retaining non‐native prosody and natural disfluencies \cite{fazel21_interspeech}. Next, we introduce a novel dynamic importance loss that builds on a recently proposed data‐weighting principle  \cite{kuo2025not} for training of our scoring model. This training loss automatically increases the contribution of synthesized instances, so as to bridge domain gaps or proficiency distribution imbalance and meanwhile curb overreliance on the generated data. Finally, the backbone of our scoring model is a Phi‐4 multimodal \cite{abouelenin2025phi} finetuned with LoRA adapters \cite{hu2022lora}, which directly fuses text prompt, as well as raw audio and ASR transcript of a spoken response, to predict the corresponding proficiency score. Our approach delivers robust performance on both seen and unseen prompts despite that only limited labeled instances are made available. Our approach not only attains superior performance on low-resource speaking assessment but also offers a replicable paradigm for other multimodal, low-resource speaking assessment tasks. In summary, our key contributions are at least three-fold:

\begin{itemize}
\item \textbf{Unified Low-resource Modeling Framework:} Combining texts generated from a large language model and their corresponding high-fidelity synthesized speech utterances to substantially expand L2 training data.
\item \textbf{Dynamic Importance Loss:} Leveraging an adaptive loss mechanism that effectively mitigates domain and distribution mismatches between synthesized and real data.
\item \textbf{Multimodal Scoring:} Developing an efficient end-to-end scoring model that fuses speech and text embeddings for robust opinion-expression scoring.
\end{itemize}

\begin{figure}[t]
  \centering
  \includegraphics[width=\linewidth]{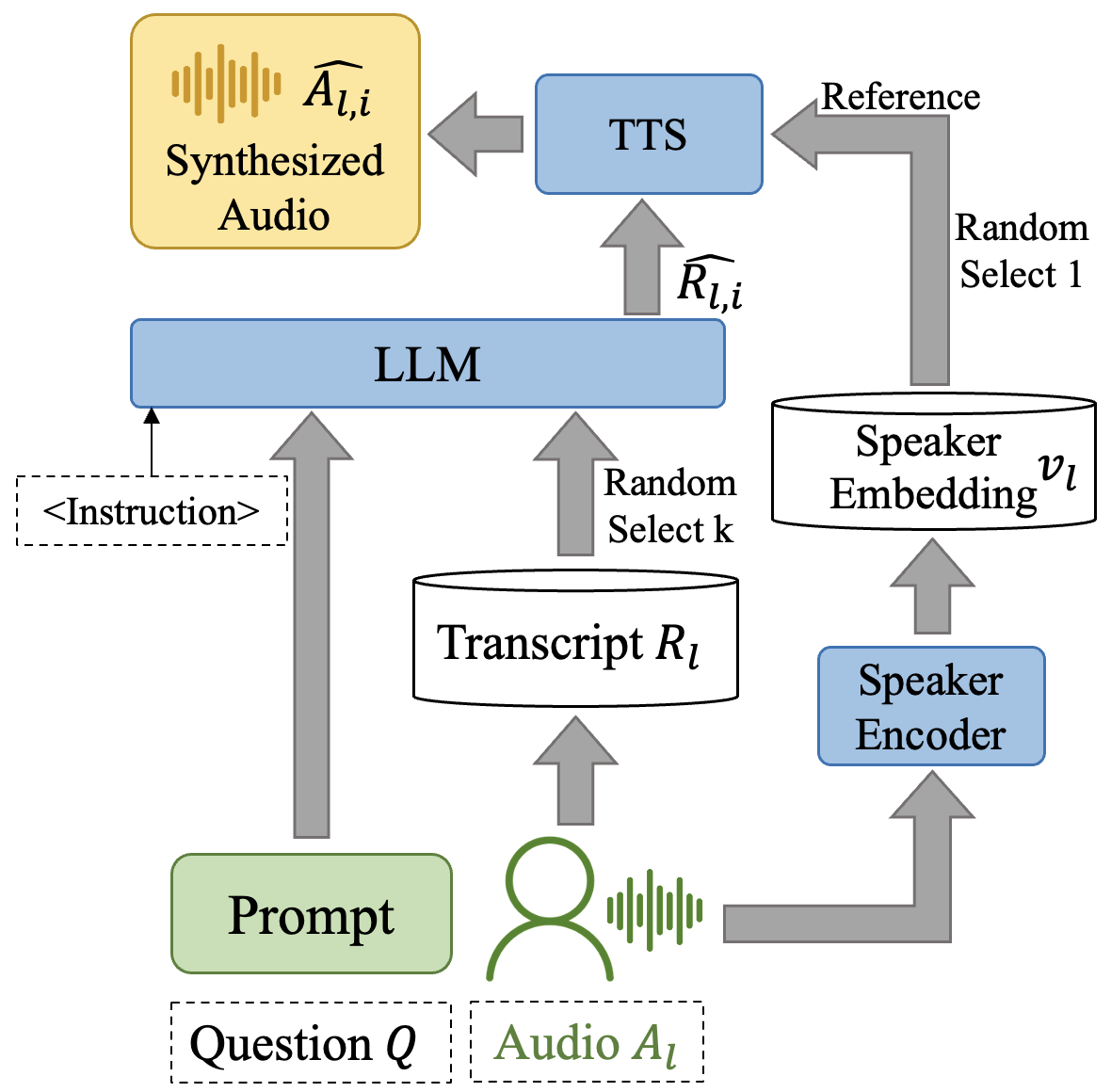}
  \caption{Pipeline for synthesized speech generation. We build in‐context prompts from $k$ real transcripts and condition a multi-speaker TTS model on LLM-generated text $\hat R_{l,i}$ and a speaker embedding $v_l$ to produce learner-like audio.}
  \label{fig:synthesized}
\end{figure}

\section{Methodology}

\begin{figure*}[t]
  \centering
  \includegraphics[width=\linewidth]{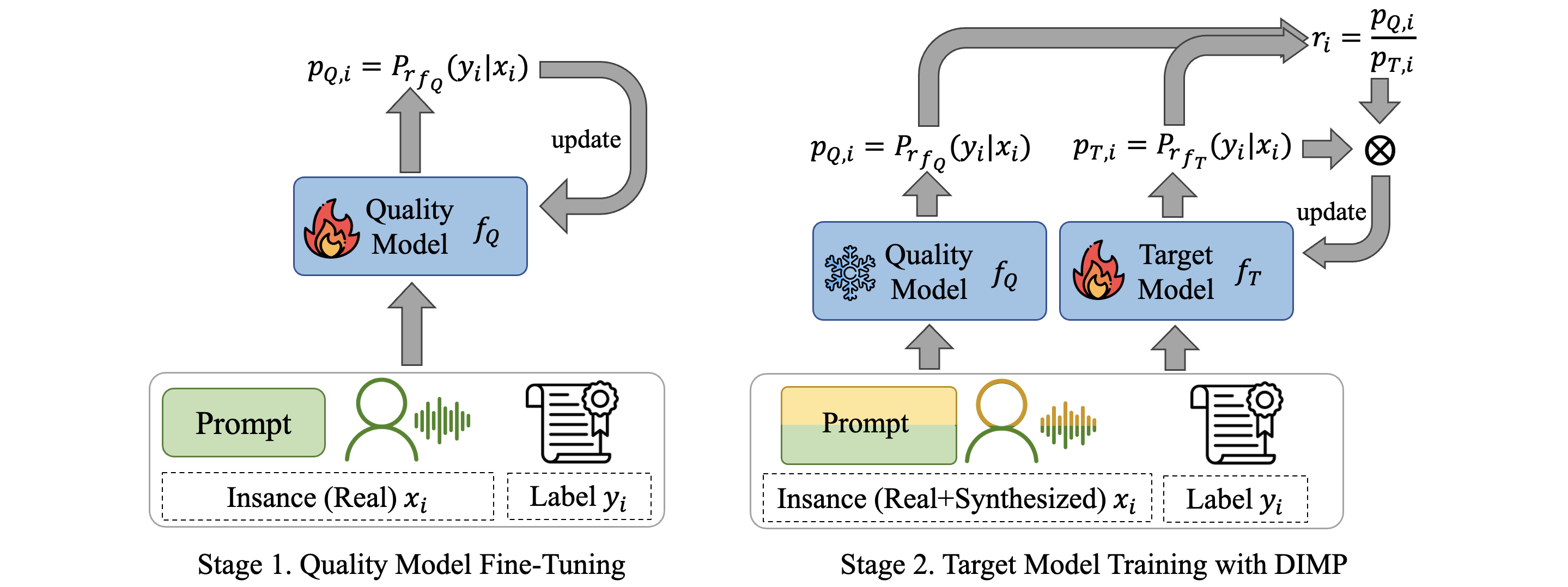}
  \caption{Two-stage training workflow. Stage 1 fine-tunes a Quality Model $f_Q$ on real data. Stage 2 trains the Target Model $f_T$ on mixed real and synthesized instances using the ratio $r_i$ and weight $w_i$ to dynamically reweight each instance’s loss.}
  \label{fig:training}
\end{figure*}

Our approach enhances opinion expression assessment in low resource settings through three tightly integrated innovations. First, a proficiency aware text synthesis module uses a large language model, prompted with randomly sampled and permuted transcripts of authentic test taker responses, to generate answers aligned with each proficiency level and faithful to real language use. Second, speaker conditioned voice cloning, driven by speaker embeddings extracted from real recordings, guides a multi speaker TTS model to convert these texts into speech while preserving non native prosody, pauses, and disfluencies; Figure \ref{fig:synthesized} illustrates the complete speech synthesis pipeline. Finally, the proposed dynamic importance reweighting loss, shown in Figure \ref{fig:training}, compares the confidences that a frozen quality model and a trainable target model assign to the correct label and then dynamically increases the weight of highly realistic synthetic samples while keeping genuine data predominant. By combining rich text augmentation, authentic speech cloning, and principled loss reweighting, our method yields a training corpus that balances diversity and authenticity and achieves robust scoring performance under strict data constraints. Implementation details and experimental results appear in the following sections.

\subsection{LLM-Based Text Synthesis}

To simulate responses from test-takers of different proficiency levels and enrich the diversity of our synthesized training instances, we leverage the in‐context learning capability of an LLM. For each target level $l$, we first randomly select $k$ transcribed responses from the genuine training dataset, denoted as $\{R_{l,1}, R_{l,2}, \dots, R_{l,k}\}$. We then concatenate the task prompt $Q$ with these examples in order to form an input prompt:

\begin{equation}
P_l = \bigl[Q;\,R_{l,1};\,R_{l,2};\dots;\,R_{l,k}\bigr],
\end{equation}
accompanied by the instruction, “Please generate your response in the style of the above examples.” During decoding, we set a relatively high temperature $\tau$ so that

\begin{equation}
\hat{R_{l,i}} \sim \mathrm{LLM}\bigl(P_l;\,\tau\bigr),\quad i=1,\dots,n,
\end{equation}
where $\hat{R_{l,i}}$ is the $i$-th synthesized response and $n$ is the number of instances generated per level. By shuffling the order of the examples and repeating this generation process across multiple rounds, we obtain a synthesized dataset of text responses that remains aligned with the prompt’s content while exhibiting rich variation in vocabulary, syntax, and viewpoints.

\subsection{Voice Cloning Synthesis}

When converting a synthesized texts $\hat{R_{l,i}}$ into the corresponding realistic speech of a test-taker, we observed that a standard TTS model tends to produce over-perfect audio which lacks natural characteristics of non-native speakers, such as disfluencies, mispronunciations, and prosodic variability, since it is typically trained on corpora of  native speakers. To remedy this, we employ a multi-speaker TTS conditioned on disparate speaker embeddings rather than a single fixed voice.

For each generated text $\hat{R_{l,i}}$, we randomly pick up a short reference audio clip $A_{l,j}$ from a real speaker at the same proficiency level $l$. A speaker encoder $E$ is then used to map this clip to an embedding:

\begin{equation}
v_{l,j} = E(A_{l,j})\,,
\end{equation}
which encodes voice identity, speaking rate, rhythm, and typical learner disfluencies. We feed both the text and embedding into the multi-speaker TTS model $G$ to produce a cloned waveform:

\begin{equation}
\hat{A_{l,i}} \sim G\bigl(\hat{R_{l,i}},\,v_{l,j}\bigr)\,,\quad i=1,\dots,n.
\end{equation}

By randomly pairing each $g_{l,i}$ with different $v_{l,j}$, we ensure the synthesized audio faithfully reproduces learner-specific traits, in contrast to the uniform, overly clean output generated by a conventional single-speaker TTS model.

\subsection{End-to-End Multimodal Training with Dynamic Importance Reweighting}

After preparing synthesized texts and cloned speech, we introduce a new dynamic importance reweighting loss that automatically adjusts the weigh of each training instance according to its fidelity to real learner behavior. This loss computes the ratio of confidence scores produced by a fixed “quality model” and by our trainable “target model,” then uses that ratio to amplify high‐quality synthesized instances and suppress poorly matched ones. At the same time, real instances receive an elevated base weight to ensure the model remains grounded in authentic speech distributions. We integrate this loss into a neural scoring model built upon the Phi-4 multimodal \cite{abouelenin2025phi} backbone and fine-tuned with LoRA adapters \cite{hu2022lora} for maximal parameter efficiency. The training phase unfolds in two stages:

\textbf{Stage 1. Quality Model Calibration}
We first fine-tune the Phi-4 multimodal \cite{abouelenin2025phi} backbone on real prompt–audio–score triplets alone, using standard cross-entropy to obtain a \emph{Quality Model} $f_Q$. This model establishes a reliable reference distribution over genuine responses of test-takers.

\textbf{Stage 2. Target Model Optimization}
Next, we train a separate \emph{Target Model} $f_T$ on the union of real and synthesized instances. For each instance $x_i$ with true label $y_i$, we compute 

\begin{equation}
p_{Q,i} \;=\;\Pr_{f_Q}(y_i\;\mid\;x_i)
\quad\text{and}\quad
p_{T,i} \;=\;\Pr_{f_T}(y_i\;\mid\;x_i),
\end{equation}
which denote the confidence that each of these two model assigns to the correct label. We then define the \emph{importance ratio}: 
\begin{equation}
r_i=\frac{p_{Q,i}}{p_{T,i}},
\end{equation}
which will be employed to amplify those synthesized instances that most closely match real-data behavior. To prevent overreliance on generated data—whose fluency often exceeds natural learner speech—we introduce a real-instance weight $\alpha>1$ with learning rate  $\eta$. The combined per-instance loss is  
\begin{equation}
\ell_i
= -\,\eta\,r_i\,w_i\,\log\bigl(p_{T,i}\bigr),
\quad
w_i = 
\begin{cases}
\alpha, & x_i\in\mathcal{D}_{\text{real}},\\
1,      & x_i\in\mathcal{D}_{\text{syn}}.
\end{cases}
\end{equation}
We optimize the average of these losses over all $N$ training instances:  
\begin{equation}
\mathcal{L}
= \frac{1}{N}\sum_{i=1}^{N}\ell_i.
\end{equation}

By introducing this novel training loss, we ensure that the augmentation of synthesized instances enriches the training set without distorting the model’s prediction capability on the authentic speech. By doing so, the estimated multimodal scoring model is anticipated to achieve robust performance under low-resource constraints.

\begin{table*}[t]
\caption{Comparison of baseline models and our approach. End‐to‐End models are marked with “\checkmark”.}
\label{tab:main_results}
\centering
\begin{tabular}{lccccc}
\toprule
Model & End‐to‐End & \multicolumn{2}{c}{Seen Test} & \multicolumn{2}{c}{Unseen Test} \\ 
      &            & Overall Acc.        & Pass–Fail  Acc.   &  Overall Acc.        & Pass–Fail  Acc.   \\ 
\midrule
Wav2vec2 \cite{baevski2020wav2vec}                & \checkmark              & 61.80\%     & 65.10\%       & 55.33\%     & 57.52\%       \\
BERT \cite{devlin2019bert}                   &               & 75.28\%     & 78.65\%       & 63.00\%     & 67.89\%       \\
Wav2vec2+BERT           &               & 73.33\%     & 77.78\%       & 64.00\%     & 67.89\%       \\
Phi‐4mm \cite{abouelenin2025phi} (only real)      & \checkmark             & 73.33\%     & 77.78\%       & 63.33\%     & 68.00\%       \\
\midrule
\textbf{Our Model}      & \checkmark             & \textbf{76.67\%} & \textbf{81.11\%} & \textbf{64.00\%} & \textbf{68.33\%} \\ 
\bottomrule
\end{tabular}
\end{table*}

\begin{figure}[t]
  \centering
  \includegraphics[width=\linewidth]{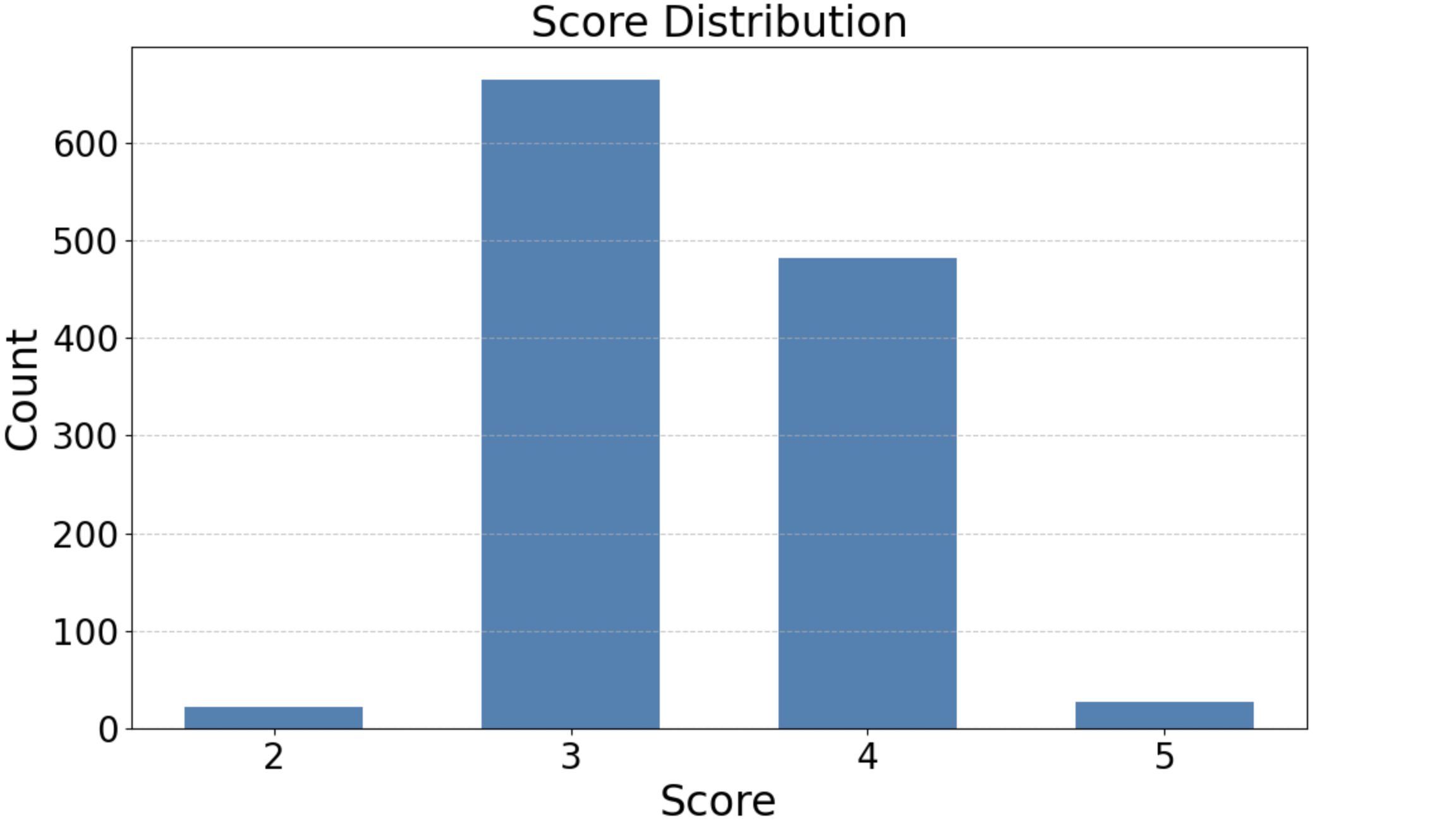}
  \caption{Score distribution for real data. }
  \label{fig:score_dis}
\end{figure}

\section{Experiments and Results}

\subsection{Dataset}
We used the LTTC GEPT Intermediate to Advanced opinion expression dataset, which contains 1 200 prompt–response–score triples evenly distributed across four prompt types within the discussion subtask. Of these, 300 triples belonging to the fourth prompt type were set aside as an unseen test set in order to simulate domain transfer. The remaining 900 triples were divided in an 8:1:1 ratio into training, validation, and seen-test sets.
Each spoken response received a 1-to-5 score from two expert raters. Following the official GEPT standard, we averaged the two scores, then applied floor rounding (no response ended with a score of 1). Scores of 4 or 5 were labeled passing and scores of 1 to 3 were labeled failing; this binary label reflects the GEPT Intermediate to Advanced speaking threshold. All spoken responses were manually transcribed for model input.
To alleviate data scarcity, we generated a synthesized corpus that matches the size and score distribution of the real training set. Figure \ref{fig:score_dis} shows the final score distributions for  real data.

\subsection{Experimental Setup}
We generate synthesized texts with OpenAI o4-mini \footnote{OpenAI o4-mini https://platform.openai.com/docs/models/o4-mini} using up to ten shuffled in-context examples per prompt and decoding at temperature $\tau=1.5$. For each text, we sample a real reference clip at the same proficiency level, extract its 512-D speaker embedding via Coqui-ai XTTSv2 \footnote{Coqui-ai XTTSv2       https://github.com/coqui-ai/TTS}, and synthesize learner-like speech with a multi-speaker TTS. Our scoring backbone is a Phi-4 multimodal \cite{abouelenin2025phi} equipped with LoRA adapters \cite{hu2022lora}. We first fine-tune a Quality Model on real prompt–speech–score triplets for three epochs using AdamW ($\beta_{1}=0.9, \beta_{2}=0.95, \epsilon=10^{-7}$), gradient clipping at $1.0$, learning rate $\alpha = 2$, batch size $16$. We then train the Target Model on the combined real and synthesized corpus under the same settings, applying a real-sample weight $\alpha = 2$ in our dynamic importance loss.

\subsection{Experiment Results}
Table \ref{tab:main_results} compares our proposed model with strong single-modality and dual-modality baselines on both the seen and unseen prompts. The baselines include an acoustics-only wav2vec2 \cite{baevski2020wav2vec} system, a text only BERT \cite{devlin2019bert} model, and a Phi‐4 multimodal \cite{abouelenin2025phi} finetuned on the real data alone.

On the test set of seen prompts, our model achieves the highest overall accuracy and pass–fail accuracy among all methods. By combining speech and text through LLM-driven augmentation and dynamic reweighting, it alleviate the downsides the systems that rely either on purely acoustic or purely textual information. It notably outperforms wav2vec in grading consistency and robustness, while surpasseing BERT despite BERT’s strong contextual encoding of transcripts. Although fine-tuning Phi‐4 multimodal model on real data alone yields competitive results, our pipeline with synthesized augmentation and the dynamic importance reweighting loss delivers the best performance, confirming the practical effectiveness of our contributions.

On the the test of unseen prompt, our  model maintains its advantage and incurs a smaller performance drop than any baseline. The synergy of diverse synthesized data and adaptive loss weighting enables more dependable scoring of novel prompts, alleviating under- or over-scoring on unfamiliar content. These results show that our unified training framework not only boosts accuracy on familiar prompts but also enhances robustness across different prompts.

\subsection{Ablation Studies}

To isolate the effects of synthesized augmentation and real‐instance weighting, we compare three variants in Table \ref{tab:ablation}: training on synthesized data alone, mixing real and synthesized with equal weight, and mixing with an elevated real‐sample weight of $\alpha=2$. Training on synthesized data by itself performs poorly, demonstrating that synthesized speech cannot substitute entirely for authentic speech. Introducing real training instances alongside synthesized ones recovers almost all of the performance lost in the first variant, validating the value of mixed‐data training. Finally, increasing the weight on real instances yields the best results on both seen and unseen prompts, confirming that our dynamic importance reweighting effectively balances synthesized diversity against fidelity to genuine speech.

\begin{table}[th]
\caption{Ablation results on the seen and unseen test sets.}
\label{tab:ablation}
\centering
\begin{tabular}{lcc}
\toprule
Variant                             & Seen Acc. & Unseen Acc. \\ 
\midrule
Only synthesized      & 42.22\%       & 46.67\%         \\
Mix ($\alpha=1$)                    & 72.22\%       & 63.67\%         \\
\textbf{Mix ($\alpha=2$)}     & \textbf{76.67\%} & \textbf{64.00\%} \\
\bottomrule
\end{tabular}
\end{table}

\section{Conclusion}
In this paper, we have proposed a unified, low-resource training framework for opinion-expression assessment that integrates LLM-driven text synthesis, speaker-conditioned voice cloning, and a novel dynamic importance reweighting loss within an end-to-end multimodal model. Empirical results on the LTTC GEPT Intermediate–Advanced dataset have shown that our approach outperforms strong single- and dual-modality baselines and generalizes robustly to both seen and unseen prompts, demonstrating the promise of generative augmentation and adaptive loss reweighting for scalable, accurate automated speaking assessment under data-scarcity scenarios.

\textbf{Limitations and Future Work}
Despite these gains, accent and hesitation mismatches persist and our experiments are limited to one language. The skewed mid-range score distribution also impairs extreme‐level discrimination. Future work will adapt the framework to additional languages and domains, address class imbalance, and explore lightweight on-device models alongside advanced augmentation strategies.

\section{Acknowledgement}
This work was supported by the Language Training and Testing
Center (LTTC), Taiwan. Any findings and implications in the
paper do not necessarily reflect those of the sponsor.

\bibliographystyle{IEEEtran}
\bibliography{mybib}

\end{document}